\RequirePackage{fix-cm}
\documentclass{article}

\usepackage{spconf,amsmath,amssymb,upgreek,graphicx,booktabs,array,float,placeins,capt-of,enumitem}
\usepackage{xurl}
\usepackage[hidelinks]{hyperref}
\usepackage{microtype}
\usepackage{flushend}

\makeatletter
\let\standardthebibliography\thebibliography
\renewcommand{\thebibliography}[1]{%
  \standardthebibliography{#1}%
  \small
  \setlength{\itemsep}{\baselineskip}%
  \setlength{\parsep}{0pt}%
  \setlength{\parskip}{0pt}%
  \clubpenalty=10000 
}
\renewcommand{\name}[1]{\gdef\@name{{\em #1}}}
\long\def\@makecaption#1#2{%
  \vskip 3pt
  \setbox\@tempboxa\hbox{#1. #2}%
  \ifdim \wd\@tempboxa >\hsize #1. #2\par
  \else \hbox to\hsize{\hfil\box\@tempboxa\hfil}%
  \fi}
\makeatother

\newcommand{\tauvoice}{$\uptau$-Voice}
\newcommand{\taumulti}{$\uptau$-Multilingual}

\title{$\uptau$-Multilingual: Benchmarking Voice Agents Across Languages}

\name{{\small Soham Ray$^{1}$\hspace{0.35em}%
Edgard dos Santos Paiva$^{1}$\hspace{0.35em}Ruben Valenzuela$^{1}$\hspace{0.35em}%
Karthik Narasimhan$^{3}$\hspace{0.35em}Keshav Dhandhania$^{1}$\hspace{0.35em}%
Victor Barres$^{2}$}}
\address{{\small $^{1}$Sierra \qquad $^{2}$Mercor \qquad
$^{3}$Princeton University}}

\begin{document}

\maketitle

\fontsize{9.5pt}{11.4pt}\selectfont

\begin{abstract}
English-only benchmarks expose only a narrow slice of voice-agent behavior. We
introduce \taumulti{}, extending \tauvoice{} to Spanish, Brazilian Portuguese,
Hindi, Korean, and Mandarin with native-speaker review and evaluation
of generated language and spoken output. Across 4,500 full-duplex
calls and five voice configurations, Spanish, Portuguese, and Hindi remain
within 3.2 task-completion points of English, but Korean and Mandarin fall by
14.7 and 8.4 points. The failure
modes also vary: Korean systems miss more
responses, Mandarin systems interrupt more often, and both struggle with tools
and entities. Grok leads task completion but scores lowest on generation
quality, motivating separate task, interaction, and generation reporting.
We release language packs, validated judges, and tools
for community-built multilingual voice-agent evaluation.
\end{abstract}


\section{Introduction}
\label{sec:introduction}

Voice agents must serve a multilingual world: English is spoken by only 19\% of
the world's population \cite{worldbank2025digital}, yet leading end-to-end voice-agent
benchmarks evaluate English alone
\cite{ray2026tauvoice,bogavelli2026evabench,lin2026fdbv3}.
Multilingual evaluation must therefore capture task completion, conversational
behavior, linguistic naturalness, and speech fidelity.

We introduce \taumulti{}, extending \tauvoice{} \cite{ray2026tauvoice} with three
contributions. First, \emph{LanguageFactory} provides a reviewed workflow for
native-speaker-authored simulators, localizing caller personas, speech rules,
and identities while preserving the underlying tasks. Second, we add
human-validated evaluation of linguistic naturalness and speech fidelity
alongside task completion and interaction measures. Third, we evaluate five
voice configurations and two text systems across six languages and three
domains.

We release the codebase \cite{taumultilingual2026release}
so the community can build callers reflecting local linguistic and cultural
norms, rather than merely translated English personas, alongside
language-specific audits of English-centric agent behavior.

\begin{figure}[!t]
\centering
\includegraphics[width=\columnwidth]{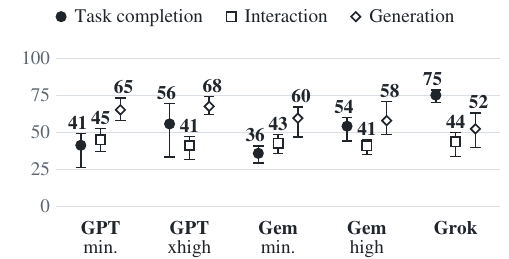}
\vspace{-1mm}
\caption{Fixed-system means with worst-to-best language ranges. Task
completion and Interaction average six languages; Generation averages the five
localized languages.}
\label{fig:model-summary}
\vspace{\baselineskip}
\captionof{table}{Related benchmarks. A check mark denotes direct evaluation and a
tilde partial coverage. Spk.: spoken; FD: full-duplex runtime; Mult.:
multilingual; IQ: interaction-quality evaluation; Lang. quality:
naturalness and language use.}
\label{tab:related}
\vspace{4pt}
\centering
\small
\setlength{\tabcolsep}{1.5pt}
\renewcommand{\arraystretch}{0.82}
\begin{tabular*}{\columnwidth}{@{\extracolsep{\fill}}lcccccc@{}}
\toprule
Work & Spk. & FD & Mult. & Task & IQ & Lang. quality \\
\midrule
\multicolumn{7}{l}{\textit{Voice-agent benchmarks}} \\
\tauvoice{} \cite{ray2026tauvoice}
  & $\checkmark$ & $\checkmark$ & -- & $\checkmark$ & $\checkmark$ & -- \\
FDB-v3 \cite{lin2026fdbv3}
  & $\checkmark$ & $\checkmark$ & -- & $\checkmark$ & $\checkmark$ & -- \\
EVA-Bench \cite{bogavelli2026evabench}
  & $\checkmark$ & $\sim$ & -- & $\checkmark$ & $\checkmark$ & $\sim$ \\
M3-DuplexBench \cite{fukuda2026m3duplexbench}
  & $\checkmark$ & $\checkmark$ & $\checkmark$ & -- & $\checkmark$ & -- \\
\addlinespace[2pt]
\multicolumn{7}{l}{\textit{Multilingual speech and dialogue}} \\
VoiceAgentBench \cite{jain2025voiceagentbench}
  & $\checkmark$ & -- & $\checkmark$ & $\sim$ & -- & -- \\
Speech-MASSIVE \cite{lee2024speechmassive}
  & $\checkmark$ & -- & $\checkmark$ & -- & -- & -- \\
X-RiSAWOZ \cite{moradshahi2023xrisawoz}
  & -- & -- & $\checkmark$ & -- & -- & -- \\
GlobalWoZ \cite{ding2022globalwoz}
  & -- & -- & $\checkmark$ & -- & -- & -- \\
\addlinespace[2pt]
\multicolumn{7}{l}{\textit{Language-quality evaluation}} \\
MQM \cite{freitag2021mqm}
  & -- & -- & $\checkmark$ & -- & -- & $\checkmark$ \\
MENLO \cite{whitehouse2026menlo}
  & -- & -- & $\checkmark$ & -- & -- & $\checkmark$ \\
\midrule
\textbf{\taumulti{}}
  & $\checkmark$ & $\checkmark$ & $\checkmark$ & $\checkmark$ & $\checkmark$ & $\checkmark$ \\
\bottomrule
\end{tabular*}
\vspace{3pt}
\end{figure}

\section{Related Work}
\label{sec:related-work}

\tauvoice{} brings the database-grounded tasks of $\uptau$-bench and
$\uptau^2$-Bench \cite{yao2024taubench,barres2025tau2bench} to full-duplex
speech, evaluating task completion and conversational dynamics but not generation
quality \cite{ray2026tauvoice}. It, EVA-Bench, and FDB-v3 are English-only
\cite{bogavelli2026evabench,lin2026fdbv3}. M3-DuplexBench covers
English and Japanese turn-taking without executed tool tasks
\cite{fukuda2026m3duplexbench}; VoiceAgentBench evaluates multilingual spoken
tool use at the query/workflow level \cite{jain2025voiceagentbench}; and
Speech-MASSIVE evaluates intent and slot prediction from recorded speech
\cite{lee2024speechmassive}.
X-RiSAWOZ and GlobalWoZ provide multilingual text dialogue with human-verified
or locale-native content \cite{moradshahi2023xrisawoz,ding2022globalwoz}; MQM
and MENLO motivate factorized, human-reviewed language evaluation
\cite{freitag2021mqm,whitehouse2026menlo}. Table~\ref{tab:related}
shows that no prior benchmark jointly evaluates multilingual full-duplex
speech, executable task success, interaction quality, and language-quality
auditing; \taumulti{} combines these capabilities with native-speaker-authored
diagnostic factors.

\section{Methodology}
\label{sec:methods}

\subsection{Matched Multilingual Benchmark}

\tauvoice{} supplies English task frames, agent policies and tools, user
scenarios, and a simulated caller \cite{ray2026tauvoice}. Its full-duplex
runtime advances in 200-ms ticks; binary rewards check task-specified
environment, action, and communication criteria. Each
language pack localizes the caller instructions, persona, voice, identity
entities, and evaluation rules while preserving the underlying tasks and
scoring. The same specification drives both voice and matched text conditions,
enabling comparisons by domain, task, system, and reasoning setting.
LanguageFactory stores this specification as a typed, versioned artifact;
Table~\ref{tab:language-factory} illustrates the Hindi pack. Cross-language
differences therefore compare complete localized systems rather than an
isolated causal effect of language.

\begin{table}[H]
\caption{LanguageFactory components illustrated with the Hindi pack.
Entity localization maps the same source identity to locale-specific backend
values without changing the task.}
\label{tab:language-factory}
\centering
\small
\setlength{\tabcolsep}{3pt}
\renewcommand{\arraystretch}{0.92}
\begin{tabular}{@{}>{\raggedright\arraybackslash}p{0.28\columnwidth}
                    >{\raggedright\arraybackslash}p{0.65\columnwidth}@{}}
\toprule
Language component & Hindi (India) \\
\midrule
Personas
& Rishika uses fast Hinglish; Imran uses patient, Urdu-inflected Hindi. \\
Speech rules
& Spell with ``R for Rajdhani''; say \emph{at} and \emph{dot} in email
addresses. \\
Task arms and entity localization
& Allison Reeves $\rightarrow$ Neha Gupta; phone: 99400 97181. \\
Language-quality factors
& Customer address (\emph{aap} vs. \emph{tum/tu}) and agreement with
\emph{aap}. \\
Review and freeze
& Pin the Rishika and Imran voices; freeze preset
\texttt{multilingual\_v1\_hindi}. \\
\bottomrule
\end{tabular}
\end{table}

Native speakers review each pack's language rules, voices, and
representative end-to-end calls.
Automated checks enforce schema, task parity, and persona assignment before
release.

\subsection{Evaluation Cohorts}

The benchmark contains 900 localized task instances: 50 tasks in each of
three domains, instantiated in six languages. Five voice configurations yield
4,500 primary calls: GPT (OpenAI \texttt{gpt-realtime-2}) at minimal and xhigh reasoning,
Gemini 3.1 Flash Live Preview at minimal and high, and Grok Voice Think Fast
1.0 \cite{openai2026gptrealtime2,google2026gemini31flashlive,xai2026grokvoicethinkfast}.
The cascaded caller uses \texttt{gpt-5.5} xhigh for text and Eleven v3 for speech
\cite{openai2026gpt55,elevenlabs2026elevenv3}. For task-completion significance
tests only, the four GPT and Gemini configurations include one additional trial.
We use the short labels GPT, Gemini, and Grok below.

The matched half-duplex text control evaluates the same 900 localized task
instances with two systems (1,800 calls): GPT-5.5
xhigh and Gemini 3.1 Pro Preview high
\cite{openai2026gpt55,google2026gemini31pro}. Because both the model and
interaction regime change, this is an end-to-end control rather than a pure
acoustic intervention.

\subsection{Retail Localization Ablations}

We evaluate GPT xhigh and Gemini high on the same 30 Hindi and Mandarin
retail tasks. The source-English identity arm restores source-English identities
while preserving the target-language conversation. The native-script database
arm replaces romanized localized names with native-script names and matched
native spell-out. Each cell has one trial per task and is reported
descriptively.

\subsection{Evaluation Measures}

Task completion is the call-level environment-reward pass rate.
\emph{Interaction} equals 100 minus the mean of five call-level failure rates:
non-response (late/missed replies or failure to yield), interruption
(talk-over), selectivity (responses to backchannels, vocal tics, or
non-directed speech), monologues over 45 seconds, and validated tool misuse
(incorrect arguments or agent-caused errors). \emph{Generation quality}
(Generation) is the share of aligned agent utterances passing naturalness and
speech-fidelity checks. Naturalness flags unnatural phrasing; fidelity uses a
multimodal Gemini 3.1 Pro Preview judge \cite{google2026gemini31pro} to compare
audio with expected text, counting severity-$\geq2$ findings or Mandarin
meaning-changing tone errors as failures. These composites are uncalibrated
summaries; language-specific diagnostics remain separate. English is excluded
only from Generation because our Naturalness rubric targets translationese and
native phrasing in the five localized languages.

We compare systems within languages and estimate matched performance gaps
relative to English for Task completion and Interaction.
We do not rank languages: each is a complete localized system, and naturalness is
not cross-language calibrated.

\subsection{Human Validation}

LLM-based judges enter results only when precision and recall exceed
.75, F1 exceeds .80, and Cohen's $\kappa>.60$. Naturalness and
language-specific diagnostics use native-language human labels. Validation
covers tool use, about 60 examples per language for
naturalness and speech fidelity, and at least 30 per diagnostic; failing
measures remain archived.

Separately, reviewers audit 30 completed calls per language, balanced across
domains and systems. They attribute the first critical error to the agent,
simulator, or infrastructure and score eight
caller-quality dimensions---including voice/prosody quality, turn-taking naturalness, and
behavioral plausibility---on a four-point scale (see the released audit rubric
\cite{taumultilingual2026release}).

\begin{figure*}[t]
\centering
\includegraphics[width=0.98\textwidth]{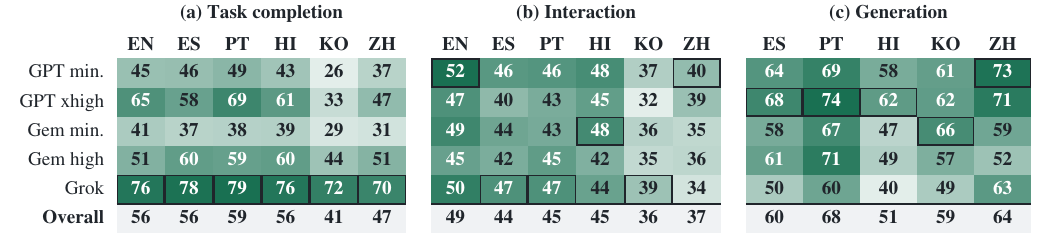}
\caption{Headline outcomes by system and language. Higher is better;
color scales normalize within panel, outlines mark language-best systems,
Overall averages the five configurations, and Generation excludes English.}
\label{fig:system-language-heatmaps}
\end{figure*}

\begin{figure}[t]
\centering
\includegraphics[width=\columnwidth]{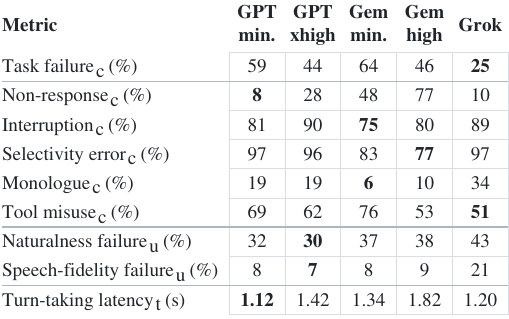}
\caption{Main-metric breakdown across languages by system; lower is better.
Bold numbers mark row minima.
Subscripts c/u/t denote call/utterance/turn; units are in parentheses.}
\label{fig:system-diagnostics}
\end{figure}

\subsection{Statistical Analysis}

Task-completion inference averages two trials per task for the four
GPT/Gemini configurations and compares each language with English over 150
matched domain--task clusters using paired permutation tests over five
contrasts and paired BCa 95\% intervals from 10,000 resamples. Grok's single
trial appears only in Figure~\ref{fig:system-language-heatmaps}(a)'s descriptive
estimates.

Interaction and Generation use 100,000 paired cluster-level label
swaps. Interaction swaps the five systems and reaggregates component
failure/opportunity counts; Generation swaps the five localized calls and
reaggregates aligned utterances. Holm correction covers five language contrasts
for Task completion and Interaction and ten system pairs for
Generation. The retail ablations are descriptive.

\section{Results}
\label{sec:results}

Figures~\ref{fig:system-language-heatmaps} and~\ref{fig:system-diagnostics}
summarize task completion, Interaction, Generation, and their component measures.
We then examine localization failures and model choice.

\subsection{Task Completion}
\label{sec:results-transfer}

Task completion is broadly preserved in Spanish, Portuguese, and Hindi but not in
Korean and Mandarin (Figure~\ref{fig:system-language-heatmaps}(a)). The
confirmatory English-relative losses for Korean and Mandarin are 18.8 points
(95\% CI: 14.8--23.3) and 9.5 points (6.0--13.4), respectively; both are
Holm-adjusted $p<.00005$, while the other three differences are not significant.
The domain pattern is similarly misaligned: telecom is hardest by
task completion (44.1\%) yet yields the highest Generation (63.2\%).

The voice--text gap favors pooled text systems in every language by
16.8--41.1 points. The smallest gap is Spanish; the largest is Korean, where
text reaches 82.0\% versus 40.9\% voice.
Because model and turn-taking also
differ, this is an end-to-end voice-system gap rather than a purely acoustic
effect.

\subsection{Interaction}
\label{sec:results-interaction}

All English-relative Interaction declines are significant: 4.6, 3.7, 3.1,
12.6, and 11.8 points for Spanish, Portuguese, Hindi, Korean, and Mandarin
(Holm-adjusted $p\leq.00055$).

Korean combines non-response (46.3\% versus 27.8\% in English) and tool-use
failure (84.8\% versus 49.8\%); Mandarin combines interruption (90.9\% versus
78.0\%) and tool-use failure (77.4\%). Selectivity failure is high throughout
(86.7--93.5\%), indicating a general weakness rather than the main language-gap
driver. Spanish and Portuguese show more monologuing than English but not the
Korean/Mandarin pattern, so similar aggregate losses require different
remedies.

\subsection{Generation Quality}
\label{sec:results-generation}

Generation produces a different model ranking: GPT xhigh leads Spanish,
Portuguese, and Hindi; Gemini minimal leads Korean; and GPT minimal leads
Mandarin (Figure~\ref{fig:system-language-heatmaps}(c)).

In their naturalness and cultural-grounding audits, native reviewers found
recurring literal or interface-like language: Spanish
\texttt{user id}, Portuguese ``nenhum voo ainda foi voado'' (no flight has yet
been flown), Korean spoken tool names such as
\texttt{check\_sim\_status}, and Mandarin \texttt{toggle data}. Targeted audits
capture different conventions: mixed formal and informal address in Portuguese;
gendered self-reference and omitted \emph{-ji} in Hindi; missing subject
honorification in Korean; and unsuitable titles and missing sentence-final
particles in Mandarin. These diagnostics contextualize Naturalness but are not
components of Generation. GPT is more natural in Hindi than Gemini and Grok
(35.1--40.6\% versus 49.7--55.5\% failure), yet GPT minimal has the most gender
errors (49.3\%). In Mandarin, GPT has nearly all modal-particle failures
(40.0--61.1\% versus 0--0.7\%).

A related, separate diagnostic is gender agreement. Agent gender was
inadvertently absent from the prompt despite using female voices, yielding
inconsistent self-grounding. Hindi
agents default to masculine self-reference (49.5\% of calls versus 5.3\% with
any feminine form). The overall gender-agreement failure rates---27.2\% in
Hindi and 29.6\% in Portuguese---combine self-reference, noun/anaphora
agreement, and caller-directed errors. Disclosure changes the mix but does not
reliably reduce the total: in 30-call-per-language ablations, failures rise from
30.0\% to 40.0\% in Hindi and fall from 30.7\% to 20.0\% in Portuguese.

Speech fidelity reveals an all-language provider pattern: 80.0\% of
Grok utterances contain no material audible error, versus 91.3--93.8\% for the
other systems. Across 72,164 scored utterances, the most common material audible
errors are audible punctuation or text-formatting artifacts (3,467 utterances),
mispronunciations (1,389), and number, date, or currency errors (1,094).

\subsection{Localization and Entity Handling}
\label{sec:results-localization}

The large Korean and Mandarin task losses concentrate in
authentication: exact lookup succeeds in 81--85\% of English calls, but only
26--56\% of Korean and 60--74\% of Mandarin calls across repeated GPT/Gemini
systems. Localized caller prompts provide the correct family-first name, yet
agents often map the spoken parts into a given-first lookup. In Mandarin, this
can break exact lookup even when callers provide the correct romanized spelling
and the agent prompt requires all tool calls, arguments, and structured outputs
in English; native-character explanations add a separate cross-script
mismatch. Telecom's
balanced split shows an exploratory crossover:
Korean task completion is 32.0\% for name+DOB versus 35.2\% for phone
authentication, whereas Mandarin is 40.8\% versus 33.6\%; method remains bundled
with task.

\begin{table}[H]
\caption{Retail entity ablations: task completion (\%) on 30 tasks.}
\label{tab:retail-ablations}
\centering
\small
\setlength{\tabcolsep}{1pt}
\renewcommand{\arraystretch}{0.86}
\begin{tabular*}{\columnwidth}{@{\extracolsep{\fill}}llrrrr@{}}
\toprule
& & \multicolumn{2}{c}{Entity} & \multicolumn{2}{c}{Entity + DB} \\
\cmidrule(lr){3-4}\cmidrule(lr){5-6}
Lang. & System & Localized & Source & Romanized & Native \\
\midrule
Hindi & GPT xhigh & 63.3 & 63.3 & 63.3 & 36.7 \\
 & Gemini high & 60.0 & 66.7 & 60.0 & 60.0 \\
Mandarin & GPT xhigh & 36.7 & 20.0 & 36.7 & 53.3 \\
 & Gemini high & 53.3 & 53.3 & 53.3 & 53.3 \\
\bottomrule
\end{tabular*}
\end{table}

In the localization ablations, source-English identities match or
outperform the localized baseline in both Hindi cells and Mandarin Gemini, but
lower Mandarin GPT by 16.7 points. Replacing romanized database values with
native-script names improves Mandarin GPT by 16.7 points and leaves Mandarin
Gemini unchanged. Hindi differs: native script leaves Gemini unchanged and
lowers GPT by 26.7 points. These 30-task cells are exploratory; exact
authentication does not ensure task completion.

\subsection{Model Choice Across Outcomes}
\label{sec:results-models}

Figure~\ref{fig:system-diagnostics} decomposes the model-level tradeoffs.
No single configuration leads every headline outcome; model choice should
follow the deployment objective:
\begin{itemize}[leftmargin=1em,labelsep=.35em,topsep=0pt,partopsep=0pt,
itemsep=1pt,parsep=0pt]
\item \textbf{Task completion:} Grok leads task completion in every language,
with a six-language mean of 75.1\%, but has the lowest Generation score
(52.3\%).
\item \textbf{Quality:} GPT xhigh leads Generation (67.5\%) at 55.7\% task
completion; GPT minimal leads Interaction (44.9\%) and has the lowest
turn-taking latency (1.12 s).
\item \textbf{Stability:} The narrowest descriptive cross-language ranges are
Grok's 8.7 points in task completion and Gemini high's 10.1 points in
Interaction. We do not interpret Generation ranges because Naturalness is not
calibrated across languages.
\item \textbf{Latency:} Higher-reasoning configurations show longer
turn-taking latency (GPT: 1.12--1.42 s; Gemini: 1.34--1.82 s). Mean call
durations are 11.7--14.2 min for GPT/Gemini versus 24.7 min for Grok.
\end{itemize}

\subsection{Validation and Robustness}
\label{sec:results-validation}

Pooled validation micro-F1 is 97.7\% for tool use, 89.4\% for naturalness,
and 90.2\% for speech fidelity; the remaining Interaction components are
deterministic.
Among measures meeting the validation thresholds, naturalness spans Hindi (F1 81.4\%,
$\kappa=.634$) to Spanish (F1 97.9\%, $\kappa=.965$), and speech fidelity spans
Korean (F1 82.4\%, $\kappa=.754$) to Spanish (F1 96.6\%, $\kappa=.955$).

The separate caller audit also supports the evaluation contract. Reviewers
attribute only 5/180 first critical errors (2.78\%) and 5/104 failed calls
(4.81\%) to the simulator, and none to infrastructure. The eight caller-quality
dimensions average 3.05/4; backchannel naturalness is lowest at 2.86.

All system pairs differ significantly on utterance Generation; the
weakest is Gemini minimal versus high (1.9 points; Holm-adjusted $p=.02378$).
Interaction effects preserve
modest Spanish/Portuguese/Hindi versus large Korean/Mandarin degradation.

\section{Limitations}
\label{sec:limitations}

We study six languages, three domains, three hosted model families, and one
speech stack. No documented open-weight model covered all five localized
languages, full-duplex speech, and live tool use; cost limited additional
languages, models, and trials.
Simulated callers may not reproduce human timing or repair. The caller
consumes agent transcripts to minimize simulator-side ASR noise, so synthesis
errors cannot affect task outcomes. GPT and
Gemini have two trials versus Grok's one, and validation sets are finite. The
frozen validation cohort includes earlier Korean and Mandarin retail runs whose
caller prompts underspecified first- and last-name roles.

\section{Conclusion}
\label{sec:conclusion}

\taumulti{} shows that multilingual voice-agent reliability extends beyond task
completion. Spanish, Portuguese, and Hindi largely match English, whereas
Korean and Mandarin lose ground through authentication, entity, and interaction
failures. Because model rankings depend on the objective, jointly reporting task
completion, interaction, and generation exposes failures hidden by English-only
or outcome-only benchmarks. We release language packs, validated judges, and
tools for community-built evaluation.

\FloatBarrier
\raggedcolsend

\section*{\centering\normalsize ACKNOWLEDGMENT}
We thank Ben Shi, Vijay Iyengar, Ola Zytek, and Ajeet Grewal for fruitful
discussions, and Clay Bavor for his continued support.

Under author supervision, OpenAI Codex and Anthropic Claude Code assisted with
targeted manuscript editing and benchmark/analysis/figure code.

\clearpage

\section{Compliance with Ethical Standards}

The benchmark uses no real customer records or recordings. Records are
synthetic and non-sensitive; calls use simulated callers and provider voices.
Reviewers authored benchmark materials and annotated outputs; they consented
and were compensated. Before release, artifacts were screened for incidental
personal data or secrets, and flagged items were excluded.

The benchmark evaluates voice-agent systems, not people or languages; its
results should not be interpreted as measures of human ability or cultural
quality.

This work was supported by Sierra and Mercor. Authors affiliated with Sierra or
Mercor are employees of their respective organizations; the authors have no
other relevant interests to disclose.

\bibliographystyle{IEEEbib}
\bibliography{references}

\end{document}